\documentclass[10pt,twocolumn,letterpaper]{article}

\usepackage{wacv}
\usepackage{times}
\usepackage{epsfig}
\usepackage{graphicx}
\usepackage{amsmath}
\usepackage{amssymb}

\usepackage{caption}
\usepackage{subcaption}
\usepackage{array,multirow}
\usepackage{threeparttable}
\usepackage{booktabs}
\usepackage{bbm}
\usepackage{xintexpr}
\usepackage[accsupp]{axessibility}

\usepackage{nicefrac}       
\usepackage{algorithm, algpseudocode}
\let\oldReturn\Return
\renewcommand{\Return}{\State\oldReturn}

\def\wacvPaperID{810} 

\wacvfinalcopy 

\ifwacvfinal

\fi

\ifwacvfinal
\usepackage[breaklinks=true,bookmarks=false]{hyperref}
\else
\usepackage[pagebackref=true,breaklinks=true,colorlinks,bookmarks=false]{hyperref}
\fi

\ifwacvfinal
\else
\fi

\begin{document}

\title{Digital and Physical-World Attacks on Remote Pulse Detection}

\author{Jeremy Speth, Nathan Vance, Patrick Flynn, Kevin W. Bowyer, Adam Czajka\\
University of Notre Dame\\
{\tt\small \{jspeth, nvance1, flynn, kwb, aczajka\}@nd.edu}
}

\maketitle
\thispagestyle{empty}

\begin{abstract}
   Remote photoplethysmography (rPPG) is a technique for estimating blood volume changes from reflected light without the need for a contact sensor. We present the first examples of presentation attacks in the digital and physical domains on rPPG from face video. Digital attacks are easily performed by adding imperceptible periodic noise to the input videos. Physical attacks are performed with illumination from visible spectrum LEDs placed in close proximity to the face, while still being difficult to perceive with the human eye. We also show that our attacks extend beyond medical applications, since the method can effectively generate a strong periodic pulse on 3D-printed face masks, which presents difficulties for pulse-based face presentation attack detection (PAD). The paper concludes with ideas for using this work to improve robustness of rPPG methods and pulse-based face PAD.
\end{abstract}
\section{Introduction}
Adversarial attacks are a security threat to computer vision systems dependent on deep learning \cite{Szegedy_ICLR_2014, Goodfellow_ICLR_2015, Su_TEC_2019}. Understanding and mitigating potential attacks is especially important for safety-critical settings such as self-driving vehicles and healthcare \cite{Kurakin_2016, Eykholt_CVPR_2018}. Much previous work focuses on adversarial attacks in object recognition, where input images are perturbed to change the predicted class. To the authors' knowledge, adversarial attacks have not been presented in tasks which produce time series signals from video, as is done in remote photoplethysmography (rPPG).

The goal of rPPG is to estimate the blood volume pulse (BVP) beneath the skin's surface from a video. Since hemoglobin absorbs a portion of the incident light, periodic fluctuations in its volume results in small color changes in the reflected light. The motivation behind rPPG is clear: we can inexpensively extract vital signs from a patient without the need for contact. In clinical settings where vital signs are used for patient assessment, accurate diagnostics are required for informed decision-making, which in turn directly influences patient outcomes. Much of the existing rPPG literature addresses challenges to estimation of accurate BVP from video, such as noise caused by motion, skin segmentation, and compression. This paper presents a new challenge in which an attacker makes subtle changes to the input video stream, either digitally or physically, in order to fool rPPG systems and generate a fake BVP.

Remote pulse detection is also a popular face presentation attack detection (PAD) idea
\cite{Li_ICPR_2016, Liu_ECCV_2016, Nowara_FG_2017, Heusch_BTAS_2018, Liu_ECCV_2018, Hernandez-Ortega_CVPRW_2018, Lin2019, Liu_WACV_2020},  and is recently applied to detect synthetically-generated face videos (``deep fakes'') \cite{Ciftci_IJCB_2020, Qi_ACM_2020, HernandezOrtega_AAAIW_2021}. Our work suggests that pulse-based face PAD methods may need to be refined to avoid adversarial attacks.

\begin{figure}
    \centering
    \includegraphics[width=\linewidth]{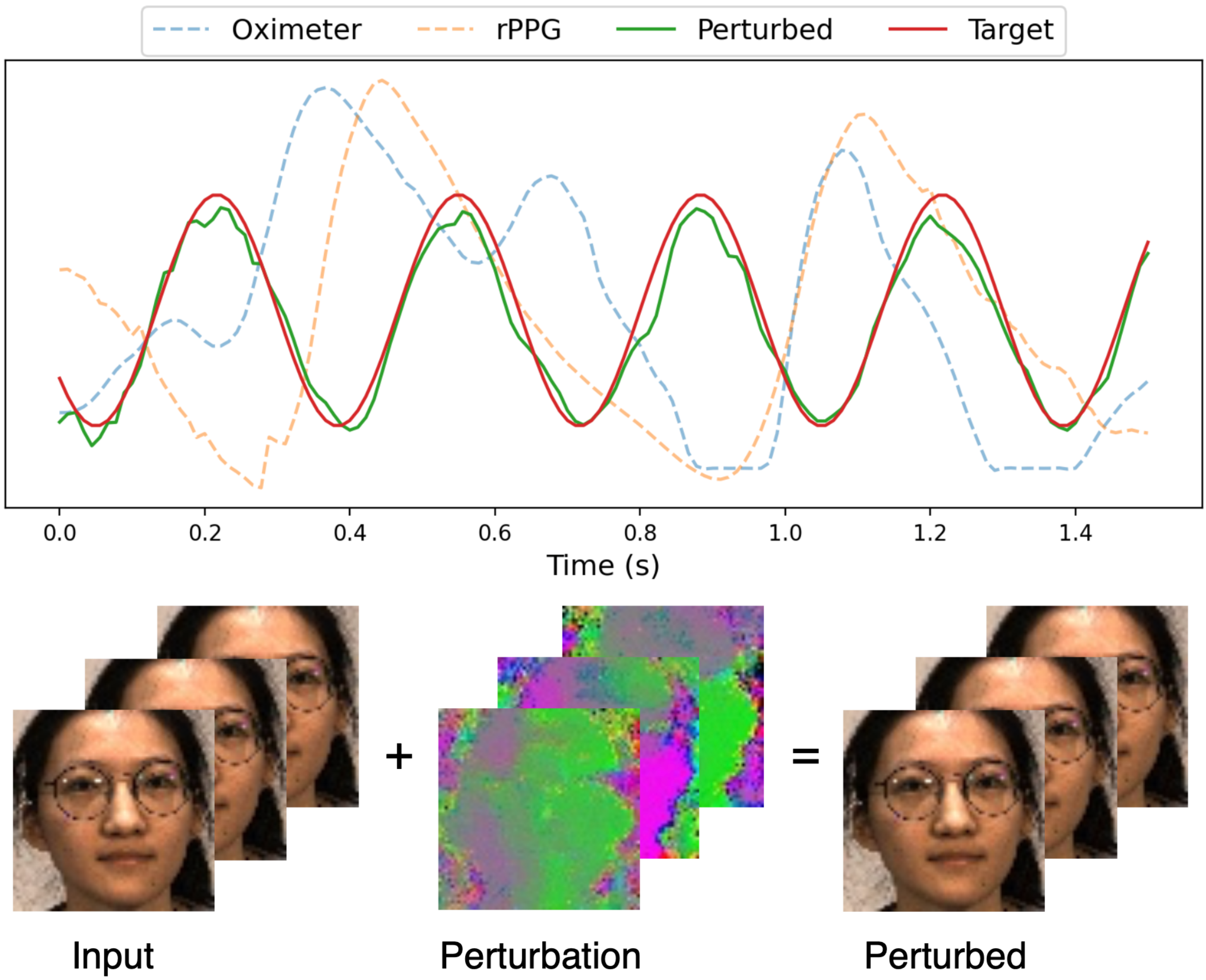}
    \caption{Input video clips are modified with low magnitude perturbations to push predictions towards a periodic target signal. Note that the perturbations are scaled to cover the full [0,255] range for visualization purposes.}
    \label{fig:teaser_figure}
\end{figure}

Since the nature of rPPG output is different than that of image classification, we specifically attack the {\it frequency} of the pulse, since heart rate is perhaps the most useful diagnostic extracted from the BVP waveforms. Figure \ref{fig:teaser_figure} shows an example of a digital attack, where the model's predictions are pushed towards a target sinusoid at a different frequency than the original BVP. This paper demonstrates methods to drive the model to predict inaccurate heart rates in the digital and physical domains.

The aforementioned adversarial attack presents a serious threat to existing pulse detectors, whether used for medical or biometric security purposes. We systematically study how such attacks could be performed digitally and physically, and briefly discuss potential techniques for detecting attacks. The contributions of this work are as follows:
\begin{enumerate}
    \item The first methodology of successful {\it digital} adversarial attacks on rPPG, demonstrated on the publicly available DDPM \cite{Speth_IJCB_2021} face videos acquired from 11 subjects along with ground-truth heart beat (via pulse oximeter);\\[-3.6ex]
    \item The first methodology and demonstration of successful {\it physical} attacks on rPPG systems, demonstrated on a newly collected dataset of four subjects with their face being illuminated by an LED with periodic changes of color;\\[-3.6ex]
    \item The first attack on pulse-based face PAD by simulating a strong pulse signal on a 3D-printed face mask.
\end{enumerate}

\section{Related Work}
Our work extends past research on white box attacks on deep neural networks. We discuss the necessary background on rPPG to establish the context for the problem, followed by the landscape of security risks to deep learning methods, and approaches to attacking vision systems.

\subsection{Remote Photoplethysmography}
Remote photoplethysmography is the process of estimating the blood volume pulse by observing changes in the reflected light from the skin. Microvasculature beneath the skin's surface fills with blood, which changes reflected color due to the light absorption by hemoglobin. Color changes attributed to blood volume are subtle and may be masked by variations due to illumination and motion, making the problem difficult in practice.

Early studies in rPPG began with stationary subjects and manually selected regions of the skin~\cite{Wieringa2005, Verkruysse2008}. These approaches applied bandpass filters and smoothing to locate the pulse signal over a single color channel. An early advancement by Poh \etal~\cite{Poh2010, Poh2011} applied blind source separation through Independent Component Analysis (ICA) to the red, green, and blue color channels.

Later works combined color channels~\cite{DeHaan2013, DeHaan2014, Wang2016, Wang2017}.
The first approach, CHROM, considered the chrominance signal, which was robust to illumination changes and movement~\cite{DeHaan2013}. A later improvement relaxed assumptions on the distortion signals introduced by movement~\cite{DeHaan2014}. Another approach utilizing all color channels examined the rotation of the skin pixels' subspace~\cite{Wang2016}. Lastly, \cite{Wang2017} introduced the plane orthogonal to skin (POS) algorithm, which mathematically verified previous color space transformation models. Their approach defines and updates a projection direction for separating specular and pulse components.

The first deep learning approach~\cite{Hsu2014} trained a regression model on ICA and chrominance features. Later, models were trained on the spatial~\cite{Hsu2017, Chen2018, Niu2020} and spatiotemporal~\cite{Yu2019} dimensions of the video rather than extracted temporal features alone. Hsu \etal~\cite{Hsu2017} trained VGG-15 on images of the frequency representation to predict heart rate. Chen \etal~\cite{Chen2018} fed frame differences and raw frames to a two-stream CNN. Using the VIPL-HR dataset, \cite{Niu2020} fed spatial-temporal maps to ResNet-18 followed by a gated recurrent unit (GRU). Yu \etal~\cite{Yu2019} constructed a 3DCNN which was given video clips and minimized the negative Pearson correlation between waveforms. A later extension added enhancement and attention networks to help with compressed video~\cite{Yu_ICCV_2019}. Lee \etal~\cite{Lee_ECCV_2020} presented a transductive learner to adapt quickly to new samples. Finally, disentangled representations were used to separate non-physiological signals from the pulse signal~\cite{Niu_ECCV_2020}.

\subsection{Adversarial Attacks on Deep Neural Networks}
Several recent works have exposed potential security threats when using deep neural networks where the input is carefully perturbed by small amounts of noise to drastically change the model's predictions~\cite{Szegedy_ICLR_2014, Goodfellow_ICLR_2015, Kurakin_2016, Eykholt_CVPR_2018, Dong_CVPR_2018}. Szegedy \etal \cite{Szegedy_ICLR_2014} showed that adversarial attacks could transfer between models, exposing the possibility for black-box attacks. Gradient-based attacks were presented in~\cite{Goodfellow_ICLR_2015}, introducing the Fast Gradient Sign Method. Iterative approaches to the optimization process further improved attacks~\cite{Kurakin_2016, Dong_CVPR_2018}. Finally, the possibility of physical-world attacks was shown in \cite{Kurakin_2016, Eykholt_CVPR_2018}, where perturbations were made outside of the digital domain.

\subsection{Face Presentation Attack Detection with rPPG}
A natural use of rPPG is face presentation attack detection (PAD), or ``liveness'' \cite{Li_ICPR_2016, Liu_ECCV_2016, Nowara_FG_2017, Heusch_BTAS_2018, Liu_ECCV_2018, Hernandez-Ortega_CVPRW_2018, Lin2019, Liu_WACV_2020}. Detecting liveness moves from attempting to estimate an accurate signal to detecting whether there is a signal or not, giving a binary decision of live or spoof. By attacking rPPG algorithms, we simultaneously attack pulse-based PAD algorithms, since a synthetic pulse signal could be projected onto a spoof medium.

\section{Pulse Estimation Model and Dataset Used}
We begin by describing the approach to accurate remote pulse estimation from RGB video. We use RPNet, a temporally dilated
3D convolutional neural network as our pulse estimator \cite{Speth_CVIU_2021}. RPNet takes a video volume as input and produces a real value for every frame, corresponding to the blood volume pulse. We chose RPNet due to its low error rates for the heart rate estimation task. Additionally, we perform our experiments on the publicly available physiological monitoring dataset, Deception Detection and Remote Physiological Monitoring Dataset (DDPM) \cite{Speth_IJCB_2021, Speth_CVIU_2021}, 
upon which RPNet was trained and evaluated.

The RPNet model, $f(\cdot)$, takes as input a video clip, $X^{N \times W \times H \times C}$, where $N$ is the number of frames taken over time, $W$ is the image width, $H$ is the image height, and $C$ is the number of channels per frame. RPNet produces a pulse waveform along the temporal dimension with a value for each frame, $\hat{Y} \in \mathbb{R}^N$. To make the rPPG task easier for RPNet, we cropped the face region and estimated landmark locations from all frames in the videos using the OpenFace toolkit~\cite{Baltrusaitis2018}. From the 68 facial landmarks, we defined a bounding box in each frame with the minimum and maximum $(x,y)$ locations. We extended the crop horizontally by 5\% to ensure that the cheeks and jaw were present. The top and bottom were extended by 30\%  and 5\% of the bounding box height, respectively, to include the forehead and jaw. We further extended the shorter of the two axes to the length of the other to form a square. The cropped frames were then resized to 64$\times$64 pixels with bicubic interpolation, hence $W=H=64$ in RPNet.

During training and evaluation, RPNet is given clips of the video consisting of $N=135$ frames (1.5 seconds). For videos longer than the clip length, we make predictions in a sliding window fashion over the full video, with a stride of half the clip length. The windowed outputs are standardized and a Hann function is applied to mitigate edge effects from convolution. Each window is then overlap-added to produce the signal for the whole video~\cite{DeHaan2013}.

To calculate the heart rate, we use a sliding window of length 30 seconds with 1-frame stride. We apply a Hamming window prior to converting the signal to the frequency domain with the Fast Fourier Transform (FFT). The frequency index of the maximum spectral peak is selected as the heart rate. A five-second moving-window average filter is then applied to the resultant heart rate signal. This approach for calculating heart rate is applied to both the oximeter's waveforms and RPNet's waveform predictions.
\section{Fast Gradient Sign Attack Method}

We consider the white-box attack scenario, where an attacker has access to the fixed weight parameters of a neural network $f(\cdot)$ and seeks to find a small perturbation for input samples, such that the network's output is closer to a target provided by the attacker. That is, given the network's forward-pass as $f(X) = \hat{Y}$, where $\hat{Y}$ is an accurate estimate of the subject's pulse, an attacker adds noise $\eta$ to the input to modify the network's output to a desired target, $\Tilde{Y}$. The attacker must find a small $\eta$ in the equation $f(X+\eta) \approx \Tilde{Y}$, since a large $\eta$ distorts the input such that it is easily discovered. Goodfellow \etal~\cite{Goodfellow_ICLR_2015} showed that finding such a perturbation is possible by using the direction of the loss gradient, $\nabla_x J(X,\tilde{Y})$. In effect, the attacker ``pushes'' the image in the direction that most strongly decreases the loss for targeted attacks:
\begin{equation}
\begin{aligned}
    \hat{Y} &= f(X - \beta \, \mathrm{sign}(\nabla_x J(X,\Tilde{Y}))).\label{eq:1}\\
\end{aligned}
\end{equation}
Equation \ref{eq:1} represents a single step in the image space towards the direction of steepest descent, where $\beta$ defines the step size. Selecting $\beta$ is difficult, and it may vary depending on the model and task. To avoid selecting $\beta$, an iterative approach named I-FGSM was proposed \cite{Kurakin_2016}, where the perturbation is repeatedly adjusted with small steps.

The MI-FGSM improvement on I-FGSM added a momentum term that allowed for avoiding local minima by accumulating a velocity during the optimization \cite{Dong_CVPR_2018}. We use MI-FGSM as the basis for our attacks on remote pulse estimation. Algorithm \ref{alg:C-MIFGSM} shows the momentum term in step 6, where a decay factor $\mu$ controls the influence of previous iterations on the current update, effectively providing momentum. The gradient is normalized by the $L_1$ norm to maintain consistent scale across iterations. The step size $\alpha$ in each step is typically defined as $\epsilon/T$, where $\epsilon$ is the maximum magnitude of the perturbation and $T$ is the number of iterations. For all experiments we use $T=50$ iterations with a maximum overall perturbation of $\epsilon=1$ on the $[0,255]$ image scale, implying $\alpha=1/50$ in each iteration. We use a decay factor of $\mu = 0.9$ in all experiments.

\begin{algorithm}[t]
\small
\caption{Physically Constrained MI-FGSM}
\label{alg:C-MIFGSM}
\begin{algorithmic}[1]
    \Require A classifier $f$ with loss function $J$; a sample video clip $x$ and target waveform $y$;
    \Require The maximum size of perturbation $\epsilon$; iterations $T$ and decay factor $\mu$.
    \Ensure
    An adversarial example $x^*$ with $\|x^* - x\|_{\infty} \leq \epsilon$.
    \State $\alpha = \nicefrac{\epsilon}{T}$;
    \State $g_0 = 0$; $x_0^* = x$;
    \For {$t = 0$ to $T-1$}
        \State Input $x_t^*$ to $f$ and obtain the gradient $\nabla_{x}J(x_t^*,y)$;
        \State Average the gradient across the spatial dimensions to get a $T \times C$ matrix with
        \vspace{-2ex}
        \begin{equation}
            \label{eq:temporal_constraint}
            \tag{\bf{Temporal}}
            \nabla_x J(x_t^*, y) =
            \frac{1}{W}\frac{1}{H}\sum^W_{w=1} \sum^H_{h=1} \nabla_x J(x_t^*, y)_{w,h};
        \end{equation} 
        \vspace{-2ex}
        \State Update $g_{t+1}$ by accumulating the velocity vector in the gradient direction as 
        \vspace{-2ex}
        \begin{align*}
            g_{t+1} = \mu \cdot g_{t} +
            \frac{\nabla_{x}J(x_{t}^*,y)}{\|\nabla_{x}J(x_{t}^*,y)\|_1};
        \end{align*}
        \vspace{-2ex}
        \State Clip the gradients to be nonpositive, such that the targeted perturbation is nonnegative with
        \vspace{-2ex}
        \begin{equation}
            \label{eq:positive_constraint}
            \tag{\bf{Nonnegative}}
            g_{t+1} = \mathrm{min}(g_{t+1}, 0);
        \end{equation}
        \vspace{-3ex}
        \State Update $x_{t+1}^*$ by applying the sign gradient as 
        \vspace{-1.5ex}
        \begin{align*}
            x_{t+1}^* = x_{t}^* - \alpha\cdot\mathrm{sign}(g_{t+1});
        \end{align*}
        \vspace{-4ex}
    \EndFor
    \Return $x^* = x_T^*$.
\end{algorithmic}
\end{algorithm}

\section{Attack Scenarios with Added Constraints}
\subsection{Scenario 1: Unconstrained Digital Attacks}
The unconstrained attacks we perform assume the attacker has access to the input video data, model weights, and predicted pulse signal over the input video. Digital attacks are significantly easier to perform than physical attacks, since the attacker may perturb every input pixel of the video volume to maximize the error in the network's prediction. Our unconstrained attack allows for the red, green, and blue channel of every pixel in every frame to be independently perturbed up to the maximum value of $\epsilon=1$ in either direction -- allowing positive or negative noise.
This is akin to targeted MI-FGSM, where we define the target signals as sine waves with various frequencies depending on the goal of the attack. Algorithm \ref{alg:C-MIFGSM} shows the steps for an unconstrained targeted attack if steps 5 and 7 are ignored.

\subsection{Scenario 2: Constrained Digital Attacks}
 In the unconstrained attack scenario, the attacker can add or subtract color values of every pixel in every frame, and can predict the unperturbed true pulse.
 This presents three key difficulties that are not necessarily feasible in the physical scenario. We thus additionally consider the {\it constrained} digital attack scenario that brings us closer to physical attacks discussed later (Scenario 3 in Sec. \ref{sec:PhysicalAttacks}). We address all of the necessary constraints independently, namely temporal (T), nonnegative (NN), and general (G), which together define our new attack approach: Constrained Momentum Iterative Fast Gradient Sign Method (C-MI-FGSM).
 
 \vspace{-1.3ex}
\subsubsection{Temporal}
In unconstrained digital attacks, noise is targeted spatially and temporally, allowing perturbations to each pixel. In a physical attack, this is difficult without a sophisticated light source. To accommodate more basic light sources that emit the same signal in all directions, we need to limit the spatial variance of the attack. Fortunately, many previous approaches to rPPG \cite{DeHaan2013, Wang2017} perform spatial averaging over the skin pixels to produce a $N\times C$ matrix for further signal extraction, implying the spatial variance is not critical.

We use this finding to constrain the added noise to the temporal and channel dimensions. In the physical system this assumes that the light is added uniformly across the scene in each frame. We achieve this by averaging the gradient over the rows and columns of each frame of the input volume, giving us a single value along each channel for every frame in the video. Line 5 of Alg. \ref{alg:C-MIFGSM} shows the aforementioned equation for temporally constraining the attack.

 \vspace{-1.3ex}
\subsubsection{Nonnegative}
Digital attacks allow the perturbation to be positive or negative. However, in the physical scenario, light cannot be subtracted from the subject's face with a basic attack device. A simple constraint is to restrict our perturbations to be nonnegative. To accomplish this we clip all normalized gradient values to $[-\infty, 0]$ in each iteration of MI-FGSM. This effectively restricts the step in each iteration to only modify the noise in frames where an increased value lowers the error between the prediction and target signal. Line 7 of Alg. \ref{alg:C-MIFGSM} shows where and how we clip the gradients.

 \vspace{-1.3ex}
\subsubsection{General}

\begin{figure}
    \centering
    \includegraphics[width=\linewidth]{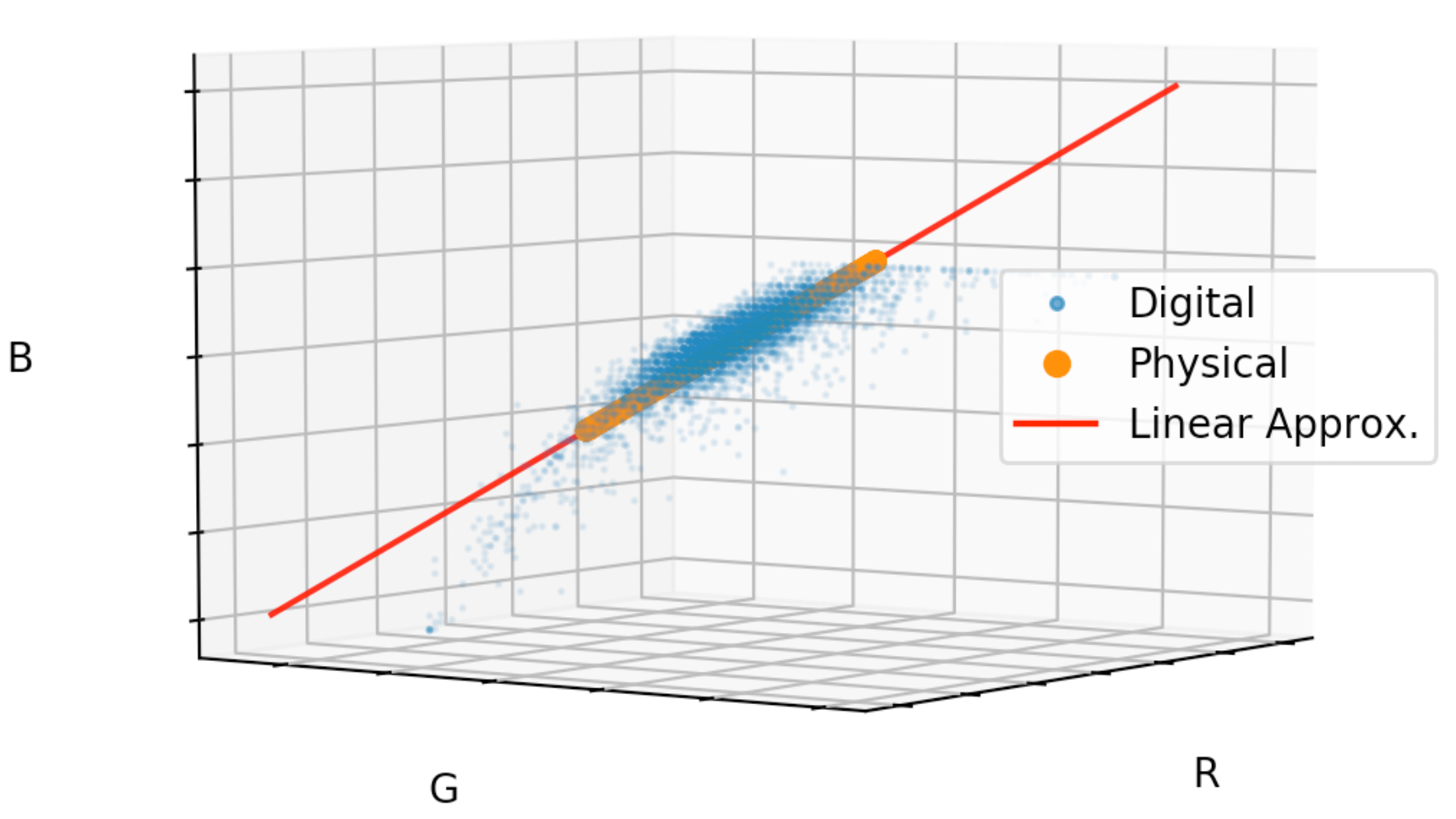}
    \includegraphics[width=\linewidth]{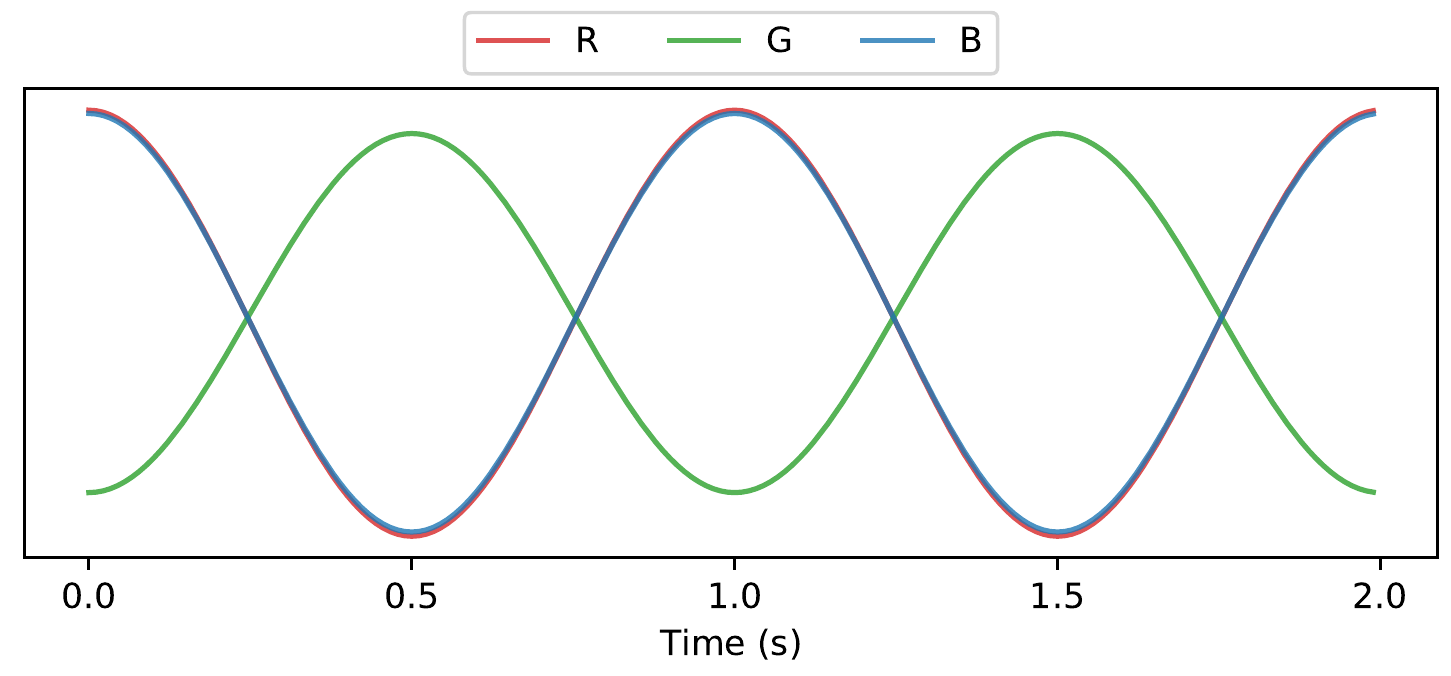}
    \caption{Top: The direction of the general physical attack is linearly approximated from digital perturbations. Bottom: Target sinusoid projected by the RGB LED along the linear estimate for a target frequency of 60 beats per minute.}
    \label{fig:physical_rgb_orbit}
    \vspace{-2ex}
\end{figure}

In the digital case, the unperturbed pulse can be predicted to define the current perturbation, meaning the attack is not necessarily independent of phase and frequency of the underlying waveform. In a physical scenario, estimating the pulse while simultaneously controlling the light pattern is substantially more computationally expensive, and small devices may not be capable of the task. We attempt to perform a general attack in which the subject's current pulse is unavailable. This is the most general attack of all, since it is agnostic to subject, heart rate, and phase.

We implement our general attack by changing the color along a vector in the RGB space, where the velocity along that vector dictates the target frequency of the attack. We observed that a simple linear approximation of data points gives acceptable goodness of fit. To find this line, we examine the per-frame perturbations from the digital attacks on the validation set with both temporal and nonnegative constraints. The top row of Fig.~\ref{fig:physical_rgb_orbit} shows a set of sampled perturbation points in the RGB space and a linear fit. 

Next, we project all data points onto the line, then remove outliers by bounding the line by 2 standard deviations of the mean. Finally, the periodic perturbations are created by projecting a sinusoid with the target frequency onto the bounded linear fit, as shown in the bottom row of Fig.~\ref{fig:physical_rgb_orbit}. This approach gives us a periodic signal in the RGB space that we can apply to any input sample.

\subsection{Scenario 3: Physical Attacks}
\label{sec:PhysicalAttacks}
A physical attack assumes the attacker has knowledge of the model weights, but no access to the input video volume in a digital format. We consider the case where the attacker physically injects noise into the scene with a targeted light source such that the 
estimated pulse is pushed towards a desired target waveform. The attacker does not have simultaneous access to the model's outputs at the time of attack, so the perturbation must be general enough for any subject at any ground truth heart rate. Our aforementioned constraints allow us to physically attack pulse estimators by adding particular periodic signals of light onto a subject's face. 

\begin{figure}
    \centering
    \includegraphics[width=\linewidth]{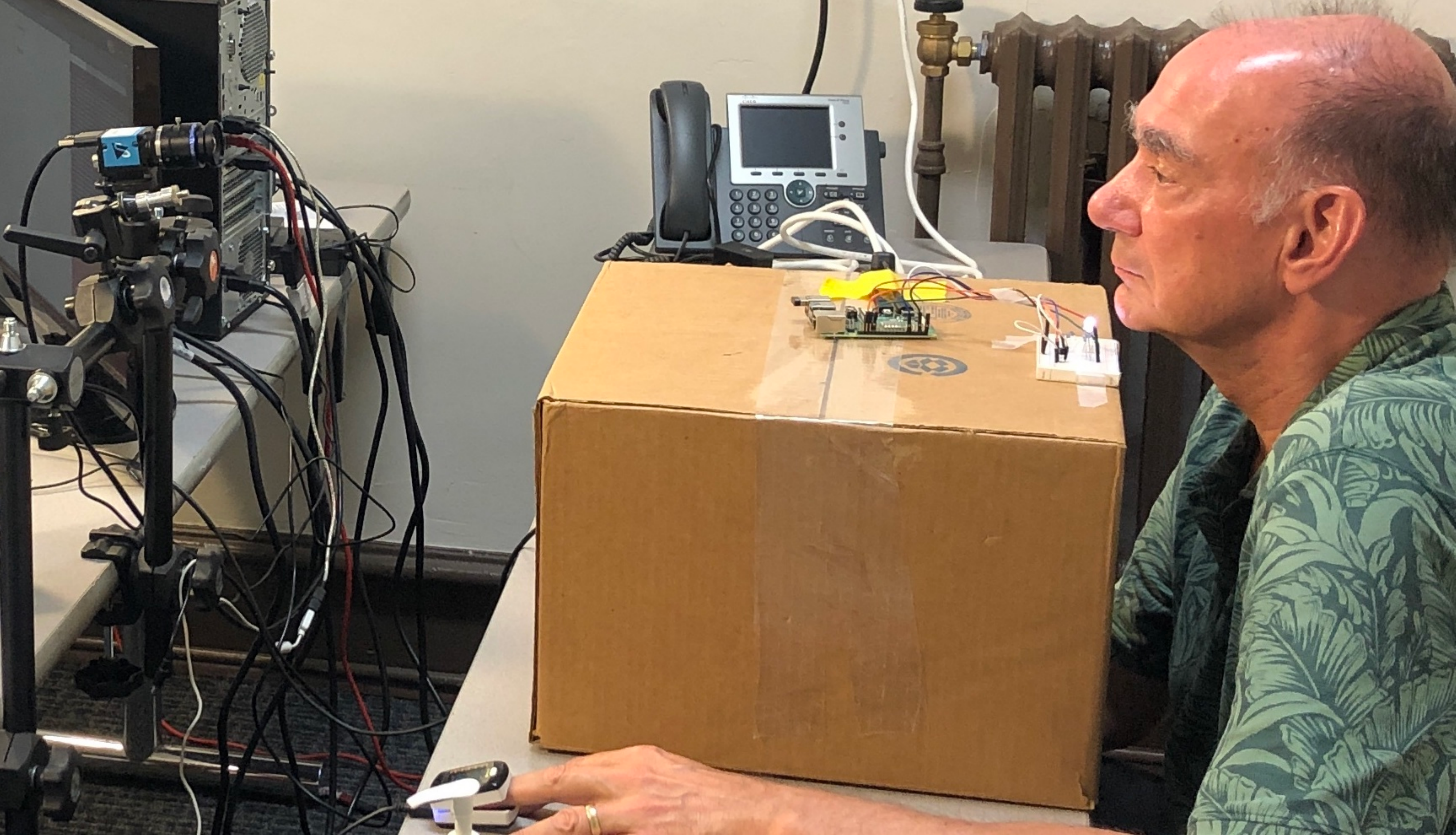}
    \caption[]{Image of the collection setting during a physical attack. The attack device is placed just below the subject's chin out of the camera's frame.}
    \label{fig:collection}
    \vspace{-2ex}
\end{figure}

\begin{figure}
    \centering
    \includegraphics[width=\linewidth]{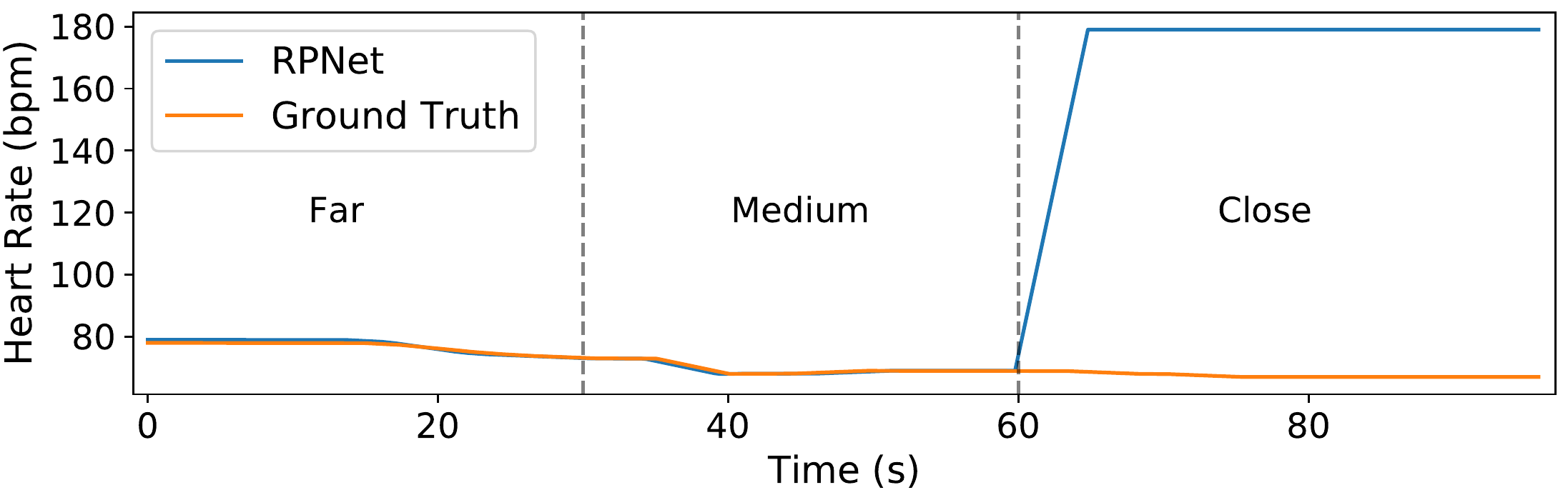}
    \caption{Ground truth and predicted heart rate for a targeted physical attack at 180 bpm. The LED was brought closer to the face every 30 seconds, where it became close enough for a successful attack at the 60 second mark.}
    \label{fig:attack_by_distance}
    \vspace{-2ex}
\end{figure}
\section{Experiments}
\subsection{Digital Attacks -- Scenarios 1 and 2}
Our experiments strive to assess the robustness of RPNet to frequency attacks. We use the DDPM test set \cite{Speth_IJCB_2021}, 
which contains 11 subjects, to evaluate the efficacy of the attacks, since RPNet was already fit to the training set. We define 12 targets as sine waves with frequencies increasing in 20 beats per minute (bpm) increments from 20 to 240 bpm.

To assess the effects of each of the individual constraints added to MI-FGSM, we perform a full ablation study. As a naming convention (seen in Tables \ref{tab:gt_errors} and \ref{tab:target_errors}), the constraint row expresses the dimensions placed on the attack. 

We examine two sets of errors in all forthcoming experiments. To analyze the efficacy of the attack, we calculate the error between the predicted heart rate of RPNet and the target heart rate of the attack. Next, we analyze the effect such an attack would have on collecting accurate diagnostics by calculating the error between the predicted heart rate of RPNet and the ground truth heart rate from the oximeter. Our chosen metrics for evaluation are mean absolute error (MAE), root mean square error (RMSE), and success rate. MAE and RMSE evaluate difference in heart rate values. 
Success rate is the percentage of time when the predicted heart rate is closer to the target than the ground truth:
\begin{equation*}
    \begin{tabular}{@{}c@{}}Success \\ Rate\end{tabular} = 100 \times \frac{1}{N}\sum_{i=1}^N{\left(|\hat{Y}_i - Y_i| > |\hat{Y}_i - \Tilde{Y}_i|\right)_{\mathbbm{1}}},
\end{equation*}
where $N$ is the number of frames for all videos in the test set, $\hat{Y}$ is the prediction, $Y$ is the ground truth, $\Tilde{Y}$ is the target, and $(\cdot)_\mathbbm{1}$ is an indicator function.

\setlength{\tabcolsep}{2pt}
\begin{table}\footnotesize
    \begin{center}
    \caption{Errors between RPNet's predicted heart rate and the ground truth during  digital and physical attacks. T indicates temporal constraint, NN indicates nonnegativity, and G indicates the general attack.}
    \label{tab:gt_errors}
    \begin{tabular}{cccccccc}
        \toprule
        Attack && Dataset
        & \begin{tabular}{@{}c@{}}rPPG\\Method\end{tabular}
        & Scenario
        & Constraint
        & \begin{tabular}{@{}c@{}}MAE \\ (bpm)\end{tabular}
        & \begin{tabular}{@{}c@{}}RMSE \\ (bpm)\end{tabular} \\
        \midrule
        \multirow{5}{*}{Digital} &&
        \multirow{5}{*}{DDPM} &
        \multirow{5}{*}{RPNet}
        & 1 & none  & 67.80 & 73.26 \\
        &&&& 2& T     & 67.80 & 73.26 \\
        &&&& 2& T+G   & 72.80 & 79.87 \\
        &&&& 2& T+NN   & 67.80 & 73.26 \\
        &&&& 2& T+G+NN & 62.96 & 70.05 \\
        \midrule
        \multirow{3}{*}{Physical} &&
        \multirow{3}{*}{ARPM-Live}
        & RPNet   & 3 &       & 26.90  & 53.73  \\
        &&& CHROM & 3 & T+G+NN & 2.69   & 13.54  \\
        &&& POS   & 3 &       & 102.82 & 132.13 \\
        \midrule
        \multirow{6}{*}{None}
        && \multirow{3}{*}{DDPM} & RPNet  & -- & -- & 2.09 & 7.30  \\
        &&& CHROM & -- & -- & 3.48 & 10.37 \\
        &&& POS   & -- & -- & 3.16 & 11.19 \\
        \cmidrule{3-8}
        && \multirow{3}{*}{ARPM-Live} & RPNet  & -- & -- & 0.06 & 0.19  \\
        &&& CHROM & -- & -- & 0.16 & 0.34 \\
        &&& POS   & -- & -- & 0.07 & 0.20 \\
        \bottomrule
    \end{tabular}%
    \vspace{-2ex}
    \end{center}
\end{table}

\setlength{\tabcolsep}{2pt}
\begin{table}\footnotesize
    \begin{center}
    \caption{Same as in Table~\ref{tab:gt_errors}, except errors are between RPNet's predicted heart rate and the targetted attack frequency.}
    \label{tab:target_errors}
    \resizebox{\columnwidth}{!}{%
    \begin{tabular}{ccccccccc}
        \toprule
        Attack && Dataset
        & \begin{tabular}{@{}c@{}}rPPG\\Method\end{tabular}
        & Scenario
        & Constraint
        & \begin{tabular}{@{}c@{}}MAE \\ (bpm)\end{tabular}
        & \begin{tabular}{@{}c@{}}RMSE \\ (bpm)\end{tabular}
        & \begin{tabular}{@{}c@{}}Success \\ Rate\end{tabular} \\
        \midrule
        \multirow{5}{*}{Digital} &&
        \multirow{5}{*}{DDPM} &
        \multirow{5}{*}{RPNet}
        & 1& none   & 0.00 & 0.00  & 99.9\% \\
        &&&& 2& T     & 0.00 & 0.00  & 99.9\% \\
        &&&& 2& T+G   & 5.00 & 12.76 & 99.9\% \\
        &&&& 2& T+NN   & 0.00 & 0.00  & 99.9\% \\
        &&&& 2& T+G+NN & 5.06 & 6.47  & 92.9\% \\
        \midrule
        \multirow{6}{*}{Physical}
        && ARPM-Live & RPNet  & 3& \multirow{3}{*}{T+G+NN} & 84.22  & 125.02  & 29.8\% \\
        && ARPM-Live & CHROM & 3 && 108.55 & 135.64  & 5.9\%  \\
        && ARPM-Live & POS   & 3 && 8.26   & 30.90   & 93.0\% \\
        \cmidrule{3-9}
        && ARPM-Mask & RPNet  & 3 & \multirow{3}{*}{T+G+NN} & 0.44  & 0.68   & 100.0\% \\
        && ARPM-Mask & CHROM & 3 && 46.08 & 92.92  & 75.0\%  \\
        && ARPM-Mask & POS   & 3 && 0.43  & 0.68   & 99.2\%  \\
        \bottomrule
    \end{tabular}%
    }
    \vspace{-2ex}
    \end{center}
\end{table}

\subsection{Physical Attacks -- Scenario 3}
Real-world digital attacks on rPPG have never been performed, so we collected a small RGB video dataset with an attack device and ground truth oximeter data. Figure~\ref{fig:collection} shows an image taken from the side of the collection process. The attack device itself is basic, only requiring commodity electronics and a programmable microcomputer (we used a Raspberry Pi) to drive an LED along the approximated line in the RGB color space. The LED was placed in close proximity to the subject's face below their chin. Note that the ambient lighting was stable, and the LED only projects the attack in a small radius. From left to right, Fig.~\ref{fig:collection} shows the camera, oximeter, and attack device below the subject's chin. The camera was oriented such that the LED was outside of the frame. 

We collected five two-minute videos from each of four subjects, to create a small (yet sufficient to support the paper claims) Adversarial Remote Physiological Monitoring (ARPM) dataset. As a control, we first collected a video from each subject without the attack. Next, for each subject we attempted attacks for frequency values in $[40, 120, 200, 300]$ beats per minute (bpm), giving us 4 attack videos and 1 unperturbed video. This set of 20 videos is denoted as ARPM-Live throughout the rest of the paper.

We found the distance between the face and LED to be an important parameter, since the luminance is relatively low from a single RGB LED compared to stable ambient lighting. Figure~\ref{fig:attack_by_distance} shows a targeted physical attack at 180 bpm, where the distance between the face and LED is decreased every 30 seconds. The clear threshold at 60 seconds illustrates the importance of the distance for successful attacks.

Along with our primary evaluation of RPNet's robustness, we examine the generalization ability of our attacks on CHROM and plane orthogonal to skin (POS). Both methods perform linear color transformations to estimate a robust pulse signal.

\begin{figure*}
    \centering
    \includegraphics[width=\linewidth]{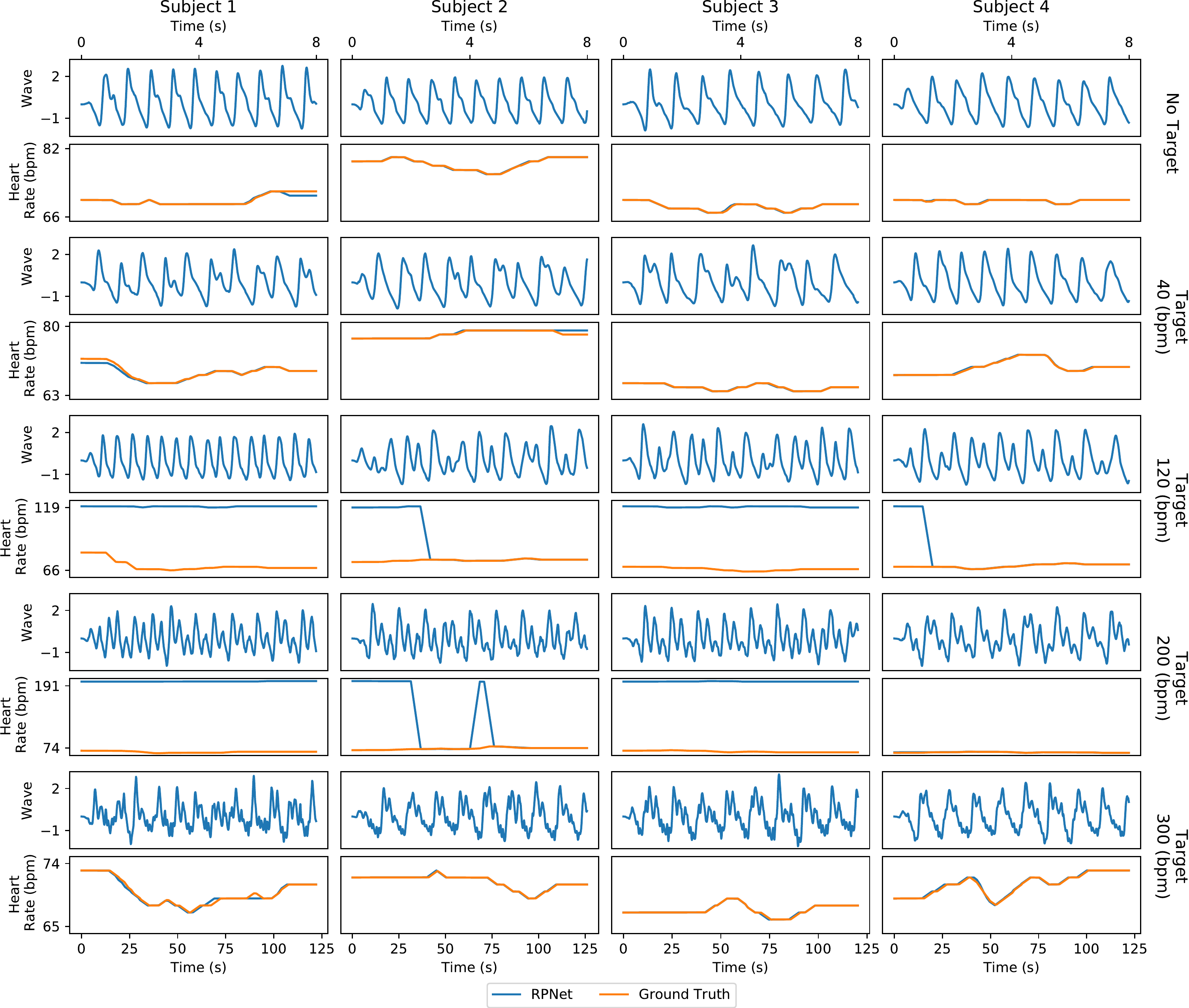}
    \caption{RPNet's predictions and the ground truth for ARPM-Live. RPNet is robust to low-frequency attacks (40 bpm), but susceptible to mid- to high-frequency attacks (120 and 200 bpm). Attacks at 300 bpm add low strength noise to RPNet's waveform predictions, but don't change the peak spectral frequency. Attacks fail or succeed in a nearly binary fashion (e.g. subject 2 at 200 bpm), where the distance likely comes into play due to our low-power LED attack device.}
    \label{fig:attack_predictions}
    \vspace{-1.5ex}
\end{figure*}

\section{Results}

\subsection{Digital Attacks -- Scenarios 1 and 2}

The top of Table \ref{tab:target_errors} shows the errors between the target and predicted heart rates. Amazingly, the MAE and RMSE for unconstrained attacks (Scenario 1) with a perturbation bound of $\epsilon=1$ is nearly 0 bpm. Consequently, the attack success rate is nearly 100\%. Since our $\epsilon$ is so small, the risk for detecting an attack upon manual inspection is low. This is somewhat unsurprising, since the pulse signal is not visible to the human eye in the first place.

We find that the added constraints still allow for extremely effective digital attacks. The errors with the temporal (T) and temporal with nonnegative perturbations (T+NN) are identical to the unconstrained attacks. The errors when using general attacks with the temporal constraint are slightly increased with a MAE of 5 bpm and RMSE of nearly 13 bpm, however the success rate is still nearly 100\%, meaning there are only a few time points where the prediction is very far from the target.

The most constrained attack (T+G+NN) gives slightly higher MAE with success rate decreased to 92.9\%. Although it is less effective than less constrained methods, it is still an achievement to perform an attack without any information on the phase, frequency, or subject being attacked.

As baseline RPNet performance, the bottom of Table \ref{tab:gt_errors} shows the errors between the ground truth and predicted heart rates without perturbations on DDPM and ARPM-Live. The MAE for DDPM is slightly more than 2 bpm, which is much higher than the 0.06 bpm on ARPM-Live, since there is minimal facial movement in ARPM sample.

The top rows of Table \ref{tab:gt_errors} show the digital attack effects on predicting the ground truth heart rate. In general, we see that unconstrained scenario 1, and scenario 2 with T and T+NN constraints give identical errors, yielding an extremely high error of 67.8 bpm. Adding the general constraint to the temporal constraint actually increases the error, which seems to indicate there are a few predictions that are very far from both the ground truth and target frequency. Finally, the most general attack gives a slightly lower error, although an MAE of nearly 63 bpm renders RPNet useless during digital attacks. \textbf{Our constrained digital attacks pushed RPNet's MAE from well within FDA approval range (with an error of 0.06 bpm) to nearly 63 bpm.}

\subsection{Physical Attacks -- Scenario 3}
Similarly to the digital experiments, we analyze the error between the ground truth oximeter and predicted heart rate, and error between target and predicted heart rate. The bottom row of Table \ref{tab:target_errors} shows the efficacy of our physical attacks from the perspective of the attacker. We find that the error between the target heart rate and predicted heart rate is quite high at 84.22 bpm, that is, our attacks are not able to consistently push the model's predictions close to the target. One cause of such high error is the target heart rate of 300 bpm, which is outside the normal physiological range, and was meant to test the possible attack frequencies. Nearly 30\% of the time our attack pushes RPNet's predictions closer to the target than the ground truth bpm.

The middle of Table \ref{tab:gt_errors} shows the effects of physical attacks from the perspective of the ground truth. We find that our physical attacks are effective in damaging predictions with respect to the ground truth. Figure \ref{fig:attack_predictions} shows RPNet's predictions and the ground truth heart rate for the entire ARPM-Live dataset. \textbf{While the physical attack device is projecting light onto the subject's face, the MAE between RPNet's predictions and the oximeter is 26.9 bpm, resulting in an error more than 425 times higher than without the attack.}

Lastly, we analyze the errors of the CHROM and POS rPPG estimators for the physical attacks. Both methods give accurate heart rate estimates on the baseline non-attack videos, with MAE values of 0.16 and 0.07 bpm for CHROM and POS, respectively. During physical attacks, the MAE between the predictions and ground truth significantly increases to 102.82 bpm for POS, while remaining relatively low at 2.69 bpm for CHROM.

The difference in attack robustness can potentially be explained by the bandpass filtering (cutoff frequencies of 42 bpm and 240 bpm) after the chrominance projection in the CHROM algorithm. When the filtering is removed, the MAE jumped to 86.26 bpm. If the filtering is applied as the final step, then the MAE is still 36.28 bpm. Interestingly, attacks within the passband are still partially mitigated in the original algorithm. These findings are both encouraging and worrying. We found that CHROM was somewhat robust without modifying the original algorithm. However, we found that the attack generalized across methods to POS, severely damaging its heart rate estimation.

\section{Impacts and Discussion}
Our work shows that physically attacking rPPG systems is feasible with a simple RGB LED in close proximity to the face. This raises serious security concerns for deploying systems in medical and biometric domains. From a medical perspective, adding noise to physiological signs may lead to inaccurate medical decisions and so cause harm to the patient. From a biometrics perspective, we present the first counterexample to accurate face presentation attack detection using rPPG, since our attack can be applied to attack mediums to simulate a pulse.

\begin{figure}
    \centering
    \includegraphics[width=0.82\linewidth]{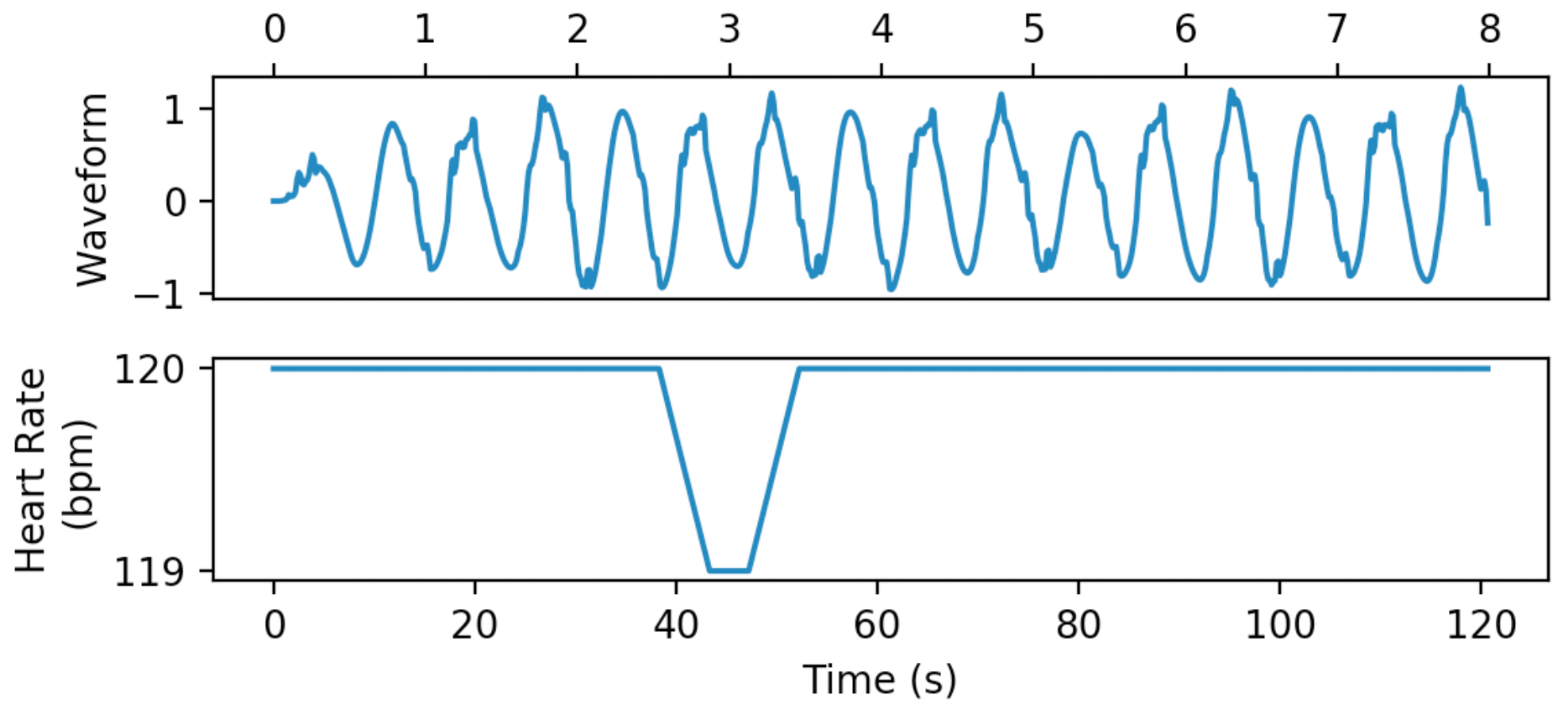}
    \raisebox{0.30\height}{\includegraphics[width=0.17\linewidth]{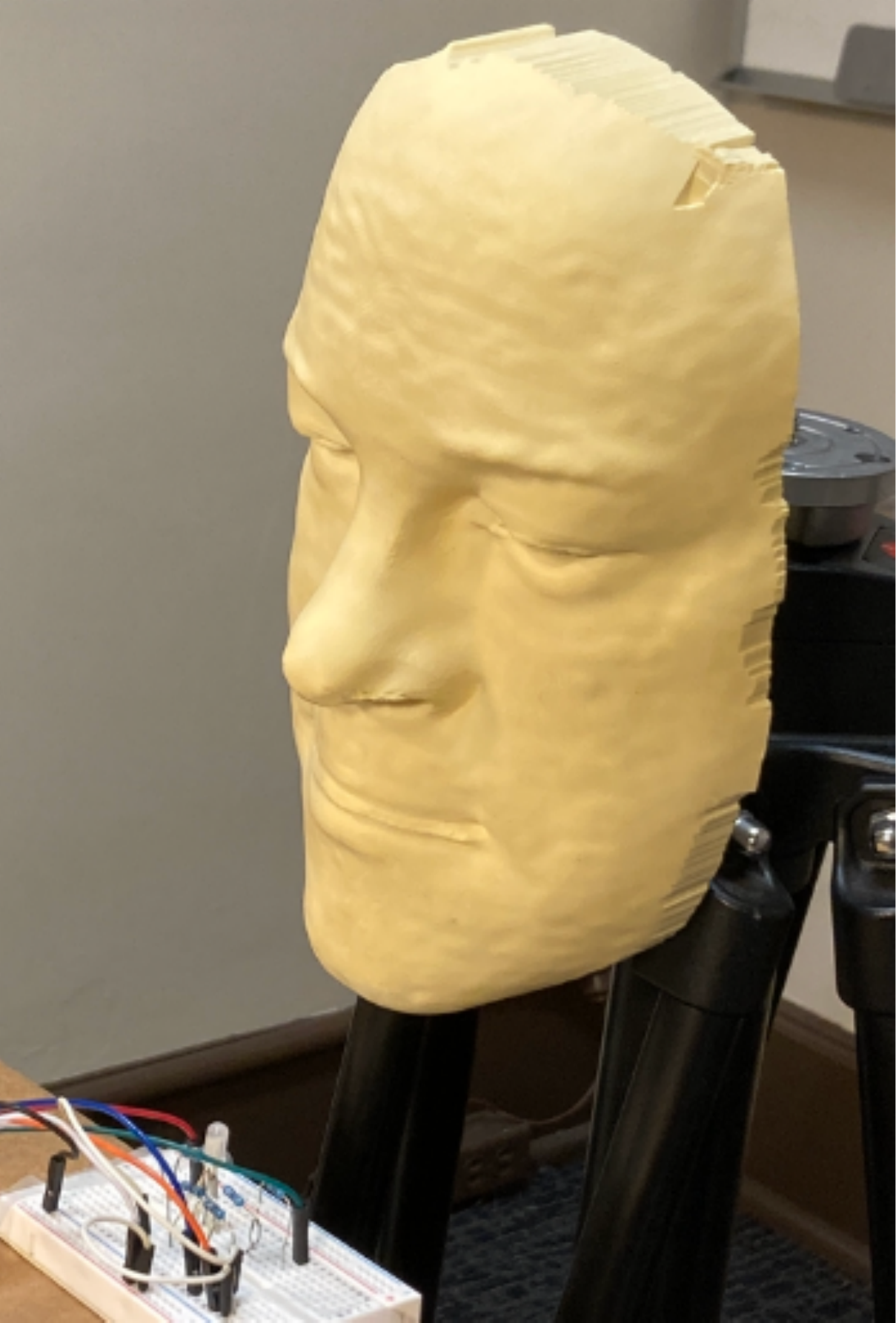}}
    \caption[]{RPNet's pulse predictions from the 3D-printed face mask (right) with a targeted attack at 120 bpm. The top row shows the predicted waveform for 8 seconds and the bottom row shows the extracted heart rate for 2 minutes.}
    \label{fig:mask_attack}
    \vspace{-2ex}
\end{figure}

To explore further the biometric security context, we project the same physical attacks onto a 3D-printed face mask to effectively simulate a live pulse signal. In the same manner as for the ARPM-Live, we collected 5 videos of the face mask with 4 attacks and 1 control to form the ARPM-Mask dataset. Figure~\ref{fig:mask_attack} shows the 3D-printed face mask along with the predicted pulse signal with a target heart rate of 120 bpm. Visually, the fake pulse signal looks quite strong, which may successfully attack rPPG-based face PAD algorithms. Although the target frequency is achieved in attacking the mask, the amplitude of the waveform is noticeably lower than for the live subjects, which could be a promising sign for pulse-based PAD methods.

We define success rate for attacking the mask as the percentage of time when RPNet's prediction is within 2 bpm of the target frequency. The bottom of Table \ref{tab:target_errors} shows the strength of our attacks. The predictions from RPNet consistently get very close to the target frequencies with an error of less than 2 bpm for 100\% of the attacks.

Identifying attack mediums and approaches early is critical towards building robust systems. One promising approach is to introduce adversarial examples while training deep learning-based rPPG models. We also found that CHROM was somewhat robust against attacks, likely due to a combination of bandpass filtering and different color projections in the RGB space. A potential mitigation strategy could leverage this finding to aggregate a pulse estimate from multiple color transformations.

Another mitigation strategy is simply using medium-wave or long-wave infrared imaging to estimate the pulse \cite{Garbey_CVPR_2004, Garbey_TBME_2007}. Our approach transmits the attack through the reflected light from the subject's face, but radiation is typically emitted in the long wave infrared spectrum \cite{Garbey_TBME_2007}. Attacking at such wavelengths would require substantially more sophisticated devices, and an understanding of the heat transfer from the face.

\section{Conclusion}
We introduced and evaluated a methodology (C-MI-FGSM) for attacking video models built for rPPG. After successfully applying C-MI-FGSM to perform digital attacks on an example deep learning-based pulse estimator, we designed a simple device and method for extending such attacks to the physical domain. We collected video from four subjects with and without physical attacks to show that our approach is capable of robust attacks without prior knowledge of the subject's current pulse. Further testing on two color-based rPPG methods showed that our physical attacks even generalize to other estimators.
In addition to the medical implications, we show that our targeted attacks can challenge face presentation attack detection. Applying our attack on a synthetic face mask results in the pulse estimator giving a strong periodic signal, presenting a problem for pulse-based liveness detection approaches. Finally, we discuss potential approaches to detecting rPPG attacks.
This paper opens a new area of research related to attacks and countermeasures for rPPG methods.

\vspace{-1ex}
\paragraph{Acknowledgements.} This research was sponsored by the Securiport Global Innovation Cell, a division of Securiport LLC. Commercial equipment is identified in this work in order to adequately specify or describe the subject matter. In no case does such identification imply recommendation or endorsement by Securiport LLC, nor does it imply that the equipment identified is necessarily the best available for this purpose. The opinions, findings, and conclusions or recommendations expressed in this publication are those of the authors and do not necessarily reflect the views of our sponsors. 

{\small
\bibliographystyle{ieee_fullname}
\bibliography{egbib}
}

\end{document}